\def\year{2020}\relax
\documentclass[letterpaper]{article} 
\usepackage{aaai20}  
\usepackage{times}  
\usepackage{helvet} 
\usepackage{courier}  
\usepackage[hyphens]{url}  
\usepackage{graphicx} 
\usepackage{graphicx}  
\usepackage{amssymb}
\usepackage{amsmath}
\usepackage{latexsym}
\usepackage{mathtools}
\usepackage{multirow}
\usepackage{xspace}

\usepackage{flowchart}
\usetikzlibrary{arrows,shapes}
\usepackage{booktabs}
\usepackage{enumitem}
\usepackage[algo2e,vlined,ruled]{algorithm2e}
\usepackage{tikz}

\usepackage{subcaption}

\DeclareMathAlphabet{\pazocal}{OMS}{zplm}{m}{n}
\usepackage{enumerate}
\usepackage{rotating}
\usepackage{newtxtext,newtxmath}
\usepackage{soul}

\usepackage{amsmath}
\usepackage{amsfonts,bm}

\def\eqref#1{equation~\ref{#1}}

\def\1{\bm{1}}

\def\vu{{\bm{u}}}

\def\vw{{\bm{w}}}
\def\vx{{\bm{x}}}
\def\vy{{\bm{y}}}
\def\vz{{\bm{z}}}

\def\m1{{\bm{1}}}
\def\mA{{\bm{A}}}

\def\mE{{\bm{E}}}

\def\mV{{\bm{V}}}
\def\mW{{\bm{W}}}
\def\mX{{\bm{X}}}
\def\mY{{\bm{Y}}}

\DeclareMathAlphabet{\mathsfit}{\encodingdefault}{\sfdefault}{m}{sl}
\SetMathAlphabet{\mathsfit}{bold}{\encodingdefault}{\sfdefault}{bx}{n}

\DeclareMathOperator*{\argmax}{arg\,max}

\usepackage{xspace}
\newcommand{\citet}[1]{\citeauthor{#1} \shortcite{#1}}
\newcommand{\citep}{\cite}

\newcommand{\Ls}{\mathcal{L}}

\newcommand{\Us}{\pazocal{U}}

\newcommand{\Ds}{\pazocal{D}}

\newcommand{\Xs}{\pazocal{X}}
\newcommand{\Ys}{\pazocal{Y}}

\newcommand{\Ni}{({\em i})~}
\newcommand{\Nii}{({\em ii})~}
\newcommand{\Niii}{({\em iii})~}

\newcommand{\ie}{{\em i.e.,}\xspace}
\newcommand{\eg}{{\em e.g.,}\xspace}

\newcommand{\Mypm}{\mathbin{\tikz [x=1.4ex,y=1.4ex,line width=.1ex] \draw (0.0,0) -- (1.0,0) (0.5,0.08) -- (0.5,0.92) (0.0,0.5) -- (1.0,0.5);}}%

\begin{document}

\title{Zero-Resource Cross-Lingual Named Entity Recognition}

\author{AAAI Press\\
Association for the Advancement of Artificial Intelligence\\
2275 East Bayshore Road, Suite 160\\
Palo Alto, California 94303\\
}

\author{M Saiful Bari$^\P$, Shafiq Joty$^\P$$^\S$ \and Prathyusha Jwalapuram$^\P$\\
  $^\P$Nanyang Technological University, Singapore \\
  $^\S$Salesforce Research Asia, Singapore \\
  \texttt{\{bari0001@e., srjoty@, jwal0001@e.\}ntu.edu.sg } \\}
  


\maketitle
\begin{abstract}

Recently, neural methods have achieved state-of-the-art (SOTA) results in Named Entity Recognition (NER) tasks for many languages without the need for manually crafted features. However, these models still require manually annotated training data, which is not available for many languages. In this paper, we propose an unsupervised cross-lingual NER model that can transfer NER  knowledge from one language to another in a completely unsupervised way without relying on any bilingual dictionary or parallel data. Our model achieves this through word-level adversarial learning and augmented fine-tuning with parameter sharing and feature augmentation. Experiments on five different languages demonstrate the effectiveness of our approach, outperforming existing models by a good margin and setting a new SOTA for each language pair.

\end{abstract}

\section{Introduction}


Named-entity recognition (NER) is a tagging task that seeks to locate and classify named entities in a text into predefined semantic types such as person, organization, location, etc. It has been a challenging problem mainly because there is not enough labeled data for most languages to learn the specific patterns for words that are part of a named entity. It is also harder to generalize from a small dataset since there can be a wide and often unconstrained variation in what constitutes names. Traditional methods relied on carefully designed orthographic features and language or domain-specific knowledge sources like gazetteers. 

With the ongoing neural tsunami, most recent approaches use deep neural networks to circumvent the expensive steps of designing informative features and constructing knowledge sources \cite{lampleNER,MaH16,Strubell:2017,DBLP:journals/corr/PetersABP17,akbik2018coling,BERT}. However, crucial to their success is the availability of large amounts of labeled training data. Unfortunately, building large labeled datasets for each new language of interest is expensive and time-consuming and we need fairly educated manpower to do the annotation. 

As many languages lack suitable corpora annotated with named entities, there have been efforts to design models for \emph{cross-lingual transfer learning}. This offers an attractive solution that allows us to leverage annotated data from a \emph{source} language (e.g., English) to recognize named entities in a \emph{target} language (e.g., German). One possible way to build such a cross-lingual NER system is to encode knowledge about the target language as constraints to regularize the training, which has been tried before for part-of-speech (POS) tagging \cite{Ganchev2010}. However, this would require extensive knowledge of the target language.

Another way is to perform cross-language projection. Most projection-based methods use a parallel sentence-aligned bilingual corpus, or a bi-text. For example, Yarowsky et al.  (\citeyear{Yarowsky:2001:IMT:1072133.1072187}) use an English NER tagger on the English side of a bi-text, then project its token-level predictions to the target side, and finally train a NER tagger on them. Wang and Manning (\citeyear{Wang-TACL-2014}) project model expectations and use them as constraints rather than directly projecting labels, to better transfer information and uncertainty across languages. Joint learning of NER tags and cross-lingual word alignments has also been proposed \cite{wang-che-manning:2013:ACL2013}. Overall, all of these methods require a bi-text with NER tags on one side, which is not typical for {low-resource} languages. Sentence-aligned parallel corpora are often not available for low-resource languages, and building such corpora could be even more expensive than building the NER dataset.


It is only recently that researchers have proposed cross-lingual NER models for low-resource languages. \citet{lin-etal-2018-multi-lingual} propose a multi-lingual multi-task architecture to develop \emph{supervised} NER models with minimal amount of labeled data in the target language. \citet{cross-ling-cmu} propose an \emph{unsupervised} transfer  model by projecting source language tags into the target language through \emph{word-to-word translation} using the unsupervised word translation model of \citet{MUSE}. However, this approach has several key limitations. First, for each target language, they need to translate from source to target and learn a brand new NER model. For this, they have to pre-compute a translation dictionary based on nearest neighbour search over the vocabulary items, which is often computationally expensive (e.g., fasttext-en-wiki has $\sim$2M items). This makes it difficult to scale time- and memory-wise. Furthermore, this often requires (as they do) a target language \emph{labeled development set} to select the best model. Therefore, although the translation process is unsupervised, their NER model is not purely unsupervised.\footnote{We use `unsupervised' to refer to cross-lingual models that do not use any NER labels in the target language.} Also, the training of the target language NER model is done without any knowledge about the source. 



\emph{Comprehensible Output} (CO) theory \cite{MERRILL95} of Second-Language Acquisition (SLA)  states ``learning takes place when a learner encounters a gap in his or her linguistic knowledge of the second language. By noticing this gap, the learner becomes aware of it and may be able to modify his output so that he learns something new about the language''. In other words, in SLA, the first language plays an important role in learning the second language.

In this paper, we propose an \textbf{unsupervised (or zero-resource)} cross-lingual neural NER model, which allows one to train a model for a target language, using labeled data from a source language. Inspired by the CO theory of SLA, we propose to learn the second language task under the supervision of the first language as opposed to completely forgetting about the first language. Thus, rather than doing {word-} or {phrase-based} translation  (Xie et al. \citeyear{cross-ling-cmu}; Mayhew et al. \citeyear{Mayhew2017CheapTF}), we choose to learn a base NER model on the source language first, and then tune the base model further in the presence of both languages to maximize the objective. 

Our framework has two encoders -- one for the source language and the other for the target. 
Our source model is based on a bidirectional LSTM-CRF \cite{lampleNER}, which we transfer to a target model in two steps. We first project the mono-lingual word embeddings to a common space through \emph{word-level} adversarial training. The word-level mapping yields initial cross-lingual links between two languages but does not take any NER information into account. Transferring task information in the cross-lingual setup is specifically challenging because languages vary in the word order. To tackle this, we propose an \emph{augmented fine-tuning} method with parameter sharing and feature augmentation, and jointly train the target model in supervision of the source model. In summary, we make the following key contributions:


\begin{itemize}[leftmargin=*]
    
    \item We propose a novel unsupervised cross-lingual NER model, assuming no labels in target language, no parallel bi-texts, no cross-lingual dictionaries, and no comparable corpora. To the best of our knowledge, we are the first to show true unsupervised results (validation by source-language) for zero-shot cross-lingual NER.

    \item {Our approach is inspired by the CO theory of how humans acquire a second language, which enables easy transfer to a new language.} Our approach only requires the tuning of the pre-trained source model on the (unlabeled) target data.

    \item We systematically analyze the effect of different {components of} the model and their contributions for transferring the NER knowledge from one language to another. 
    
    \item We report sizable improvements over state-of-the-art  cross-lingual NER methods on five language pairs encompassing languages from different families ($2.43$ for Spanish, $2.21$ for Dutch, $6.14$ for German, $7.1$ for Arabic, $5.73$ for Finnish). Our method also outperforms the models that use cross-lingual and multilingual external resources.
    
    \item We have released our code for research purposes.\footnote{\url{https://github.com/ntunlp/Zero-Shot-Cross-Lingual-NER}} 

\end{itemize}

\section{Problem Definition}

Our objective is to transfer NER knowledge from a source language (e.g., English) to a target language (e.g., German) in an unsupervised way. While doing so, we also wish to provide the landscape of the probable solutions and analyze different solution stages and the importance of different components of the neural model. We make the following assumptions.

\begin{itemize}[noitemsep]
    \item We have access to mono-lingual corpora for both source and target languages to create  pretrained word embeddings such as fasttext \cite{grave2018learning}. 
    \item For training, we assume that we have NER labels only for the source language dataset.
    \item We consider two validation scenarios for model selection: \Ni we have access to a labeled target language validation set, and \Nii only source language validation set is available. 
\end{itemize}

Learning \textbf{cross-lingual models} involves two fundamental steps: \Ni learn a mapping between the source and the target language, and \Nii retrain the mapped resources to maximize the task objective. These two steps can be done separately or jointly. For example, \cite{cross-ling-cmu} first translate the source sequences to target word-by-word (step \textit{i}), then they learn a target language NER model using the translated texts and projected NER tags (step \textit{ii}). However, as mentioned before, this approach has several key limitations. Besides, training over the (translated) source sequence  makes the sequence encoder more dependent on the source language order, which could introduce noise for the target language. 

In contrast, we propose to perform mapping and task transfer \emph{jointly}. Our model comprises two encoders -- one for the source language and the other for the target. We first train a base NER model on the source language, and use it to jointly train the target model through adversarial learning and augmented fine-tuning. This way, the model is able to learn from both source and target sequences. In the following, we first describe our base model, then we present our novel unsupervised cross-lingual transfer approach.

\section{Our Source (Base) Model} \label{subsec:base}

Our source (base) model has the same architecture as Lample et al. (\citeyear{lampleNER}), as shown in Figure \ref{fig:model} (the left portion). Given an input sentence $s = (w_1, \ldots, w_m)$ of length $m$, we first encode each token $w_k$ with a \emph{character-level} bi-LSTM \cite{hochreiter1997long}, which gives a token representation $\vw_k^{\text{ch}}$ by sequentially combining the current input character representation with the previous hidden state in both directions. The character bi-LSTM (shown at the bottom in the box) captures orthographic properties (\eg\ capitalization, prefix, suffix) of a token. For each token $w_k$, we also have a word embedding $\vw_k^{\text{wr}}$ that we fetch from a pretrained word embedding matrix. The pretrained word vectors capture distributional semantics of the words. We concatenate the character-level representation of a word with its word embedding to get the combined representation  $\vx_k = [\vw_k^{\text{ch}}; \vw_k^{\text{wr}}]$.

Let $\mX = (\vx_1, \ldots, \vx_m)$ denote the representation of the words in the sentence that we get from the character bi-LSTM and embedding lookup. $\mX$ is then fed into another \emph{word-level} bi-LSTM, which is also processed recurrently to obtain contextualised representations of the words.

The word-level bi-LSTM captures contextual information by propagating information through hidden layers, and can be used directly as a feature for NER classification. However, its modeling strength is limited compared to structured models that use global inference to model consistency in the output, especially in tasks having strong dependencies between output labels such as NER. Therefore, instead of classifying words independently with a Softmax layer, we model them jointly with a CRF layer (Lafferty et al. \citeyear{Lafferty01}).

For an input-output sequence pair $(\mX, \vy)$, we define the joint probability distribution as follows. 
\begin{eqnarray}
p(\vy|\mX) = \frac{1}{Z(\theta_s)} \prod_{i=1}^m \underbrace{\psi_n(y_i|\vu_i, \mV)}_\text{node factor} \hspace{-0.02cm} \prod_{i=0}^m \underbrace{\psi_e(y_{i,i+1}|{\mA})}_\text{edge factor} \label{eq:pcrf}
\end{eqnarray}
\normalsize
\noindent where $(\vu_1, \cdots, \vu_m)$ are the LSTM encoded contextualized word vectors, and $\psi_n(y_i=j|\vu_i, \mV) = \exp (\mV_{j}^T\vu_i)$ is the node-level score with $\mV$ being the weight matrix, $\psi_e$ is the transition matrix parameterized by $\mA$, and $Z(.)$ is the normalization constant to ensure a valid probability distribution, and $\theta_s$ denotes all the parameters of the (source) model. The cross entropy loss for the $(\mX, \vy)$ sequence pair is:
\begin{eqnarray}
\Ls_s(\theta_s) \hspace{-0.05cm}= \hspace{-0.05cm}- \hspace{-0.05cm} \sum_{i=1}^m \hspace{-0.05cm}\log \psi_n(y_i|\vu_i, \mV)  - \hspace{-0.05cm} \hspace{-0.05cm}\sum_{i=0}^m \log \mA_{i,i+1} 
  \hspace{-0.05cm}+\hspace{-0.05cm}\log Z \label{eqn:crfloss}
\end{eqnarray}
\normalsize
\noindent We use Viterbi decoding to infer the most probable tag sequence for an input sequence, $\vy*$ $=$ 
$\argmax_{\vy} p(\vy | \mX, \theta_s)$. 

Following Lample et al. (\citeyear{lampleNER}), we use a point-wise dense layer to transform the word representations before passing them to the CRF layer. As described later, the dense layer works as a \emph{common encoder} in our cross-lingual model through which the two encoders share task information and common language properties.

\section{Our Cross-Lingual Model}

{Our main goal is to learn a mapping of NER distributions between source and target languages.} Neural approaches to NER depend heavily on fixed or contextualized pretrained  embeddings \cite{ELMO,BERT,akbik2018coling}. However, when we learn the embeddings for two different languages separately, their distribution spaces are very different even for closely related languages \cite{Anders-18}. For example, Figure \ref{tsne-plot:1} shows the t-SNE plot for NER tagged monolingual embeddings for English and Spanish. We see that the distributions are very different. Mapping these two distributions is indeed a very challenging task, especially in the unsupervised setup where no parallel data or dictionary is given. The challenge is further compounded by the requirement that the mappings should also reflect NER information; the effective modeling of NER requires the consideration of sequential dependencies, which generally vary between two languages under consideration.

\begin{figure}[t!]
    \centering
    \includegraphics[scale=.55]{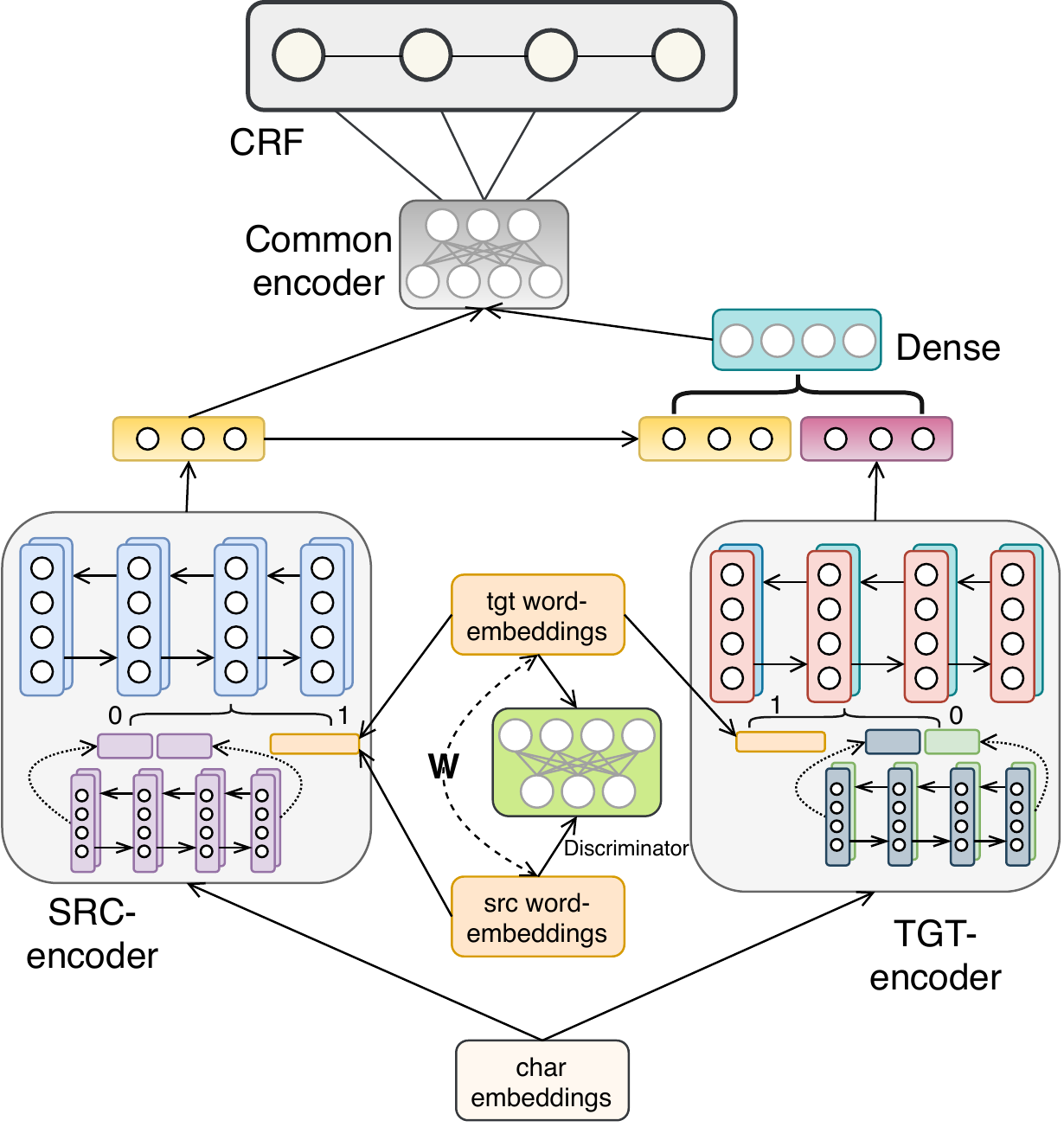}
    \caption{Our proposed model for unsupervised Cross-lingual Named Entity Recognition.}
    \label{fig:model}
\end{figure}

Figure \ref{fig:model} shows the overall architecture of our cross-lingual NER model. We add {three new components} to the base model described in the previous section: \Ni a separate encoder for the target language with shared character embeddings (box on the right) followed by a target-specific dense layer, \Nii word-level adversarial mappers that can map word embeddings from one language to another (shown in the middle of the two boxes), and \Niii an augmented fine-tuning method with parameter sharing and feature augmentation. 

\subsection{Target Encoder with Shared Character Embedding}
{Our target encoder parameterized by $\theta_t$ has the same architecture as the source encoder -- a character-level bi-LSTM followed by a word-level bi-LSTM. Having a separate encoder as opposed to a shared one allows us to explicitly model specific characteristics (e.g., morphology, word order) of the respective languages. However, this also adds an additional challenge on how to effectively share the NER knowledge between the two encoders.} 

To promote knowledge sharing through cross-lingual mapping, we share the character embeddings of the two languages by defining a common embedding matrix. If two languages share alphabets or words, these common features can be used as a prior to learn the mapping.\footnote{{We also tried subword units with BPE. However, given that the datasets are small, it did not give any additional gain.}}

\subsection{Word-level Adversarial Mapping} \label{subsec:wrd}

Sharing of character embeddings works only for languages that share alphabets. Even for languages sharing alphabets, it can only provide an initial mapping that is often not good enough to learn cross-lingual mappings. To learn the word-level mapping in an unsupervised way, we adopt the adversarial approach of Conneau et al. (\citeyear{MUSE}).

Let $\Xs = \{\vx_1, \ldots, \vx_n\}$ and $\Ys = \{\vy_1, \ldots, \vy_m\}$ 
be two sets consisting of $n$ and $m$ word embeddings of $d$-dimensions for a source and a target language, respectively. We assume that $\Xs$ and $\Ys$ are trained independently from monolingual corpora. Our aim is to learn a mapping $f(\vy)$ in an unsupervised way (\ie\ no bi-lingual dictionary is given) such that for every $\vy_i$, $f(\vy)$ corresponds to its translation in $\Xs$. Let $\mW_{t \rightarrow s}$ denote the linear mapping weight from target to source, and $\theta_D$ denote the parameters of a discriminator $D$ (a binary classifier). We define the discriminator and adversary losses as follows.
\begin{align}
\mathcal{L}_{D}(\theta_{D}|\mW_{t \rightarrow s}) = - \frac{1}{m} &\sum_{j=1}^{m} \log P_{\theta_{D}}(\text{src}=0|\mW_{t \rightarrow s} \vy_j) \nonumber\\[-0.6em]   
&- \frac{1}{n} \sum_{i=1}^n \log P_{\theta_D}(\text{src}=1|{{\vx_i}}) \\
\mathcal{L}_{\text{adv}}(\mW_{t \rightarrow s}|\theta_{D}) = - \frac{1}{m} &\sum_{i=1}^{m} \log P_{\theta_D}(\text{src}=1| \mW_{t \rightarrow s} \vy_j) \nonumber \\[-0.6em]
&-\frac{1}{n} \sum_{i=1}^n \log P_{\theta_D}(\text{src}=0|{\vx_i}) \label{adversaryAloss}
\end{align}
\normalsize

\noindent where $P_{\theta_D}(\text{src}|\vz)$ is the probability according to $D$ to distinguish whether $\vz$ is coming from the source ($\text{src}=1$) or from the target-to-source mapping ($\text{src}=0$). The mapper $\mW_{t \rightarrow s}$ is trained jointly to fool the discriminator $D$.

Adversarial training gives an initial word-level mapping, which is often not good enough. A refinement step follows, to enrich the initial mapping by considering the global properties of the embedding spaces. Following Conneau et al. (\citeyear{MUSE}), we use refinement with the Procrustes solution, where we first induce a \textbf{seed dictionary}  using the learned mapper from our adversarial training. In order to find the nearest source word ($\vx$) of a target word ($\vy$) in the common space, we use the Cross-domain Similarity Local Scaling (CSLS).
With the seed dictionary, we apply the following \textbf{Procrustes} solution to improve the initial mappings, $\mW_{t \rightarrow s}$.

\begin{align}
\mW_{t \rightarrow s} &= VU^T, \text{  where } U \Sigma V^T = \text{SVD}(\mX^T \mY)
\end{align}
\normalsize

\noindent We perform this fine-tuning iteratively: induce a new dictionary using CSLS on the newly learned mapping, then use the dictionary in the Procrustes solution to improve the mapping. The mapper for source to target $\mW_{s \rightarrow t}$ can be similarly trained to map the source embedddings to the target space.

\begin{figure}[t!]
    \centering
  \includegraphics[scale=.4]{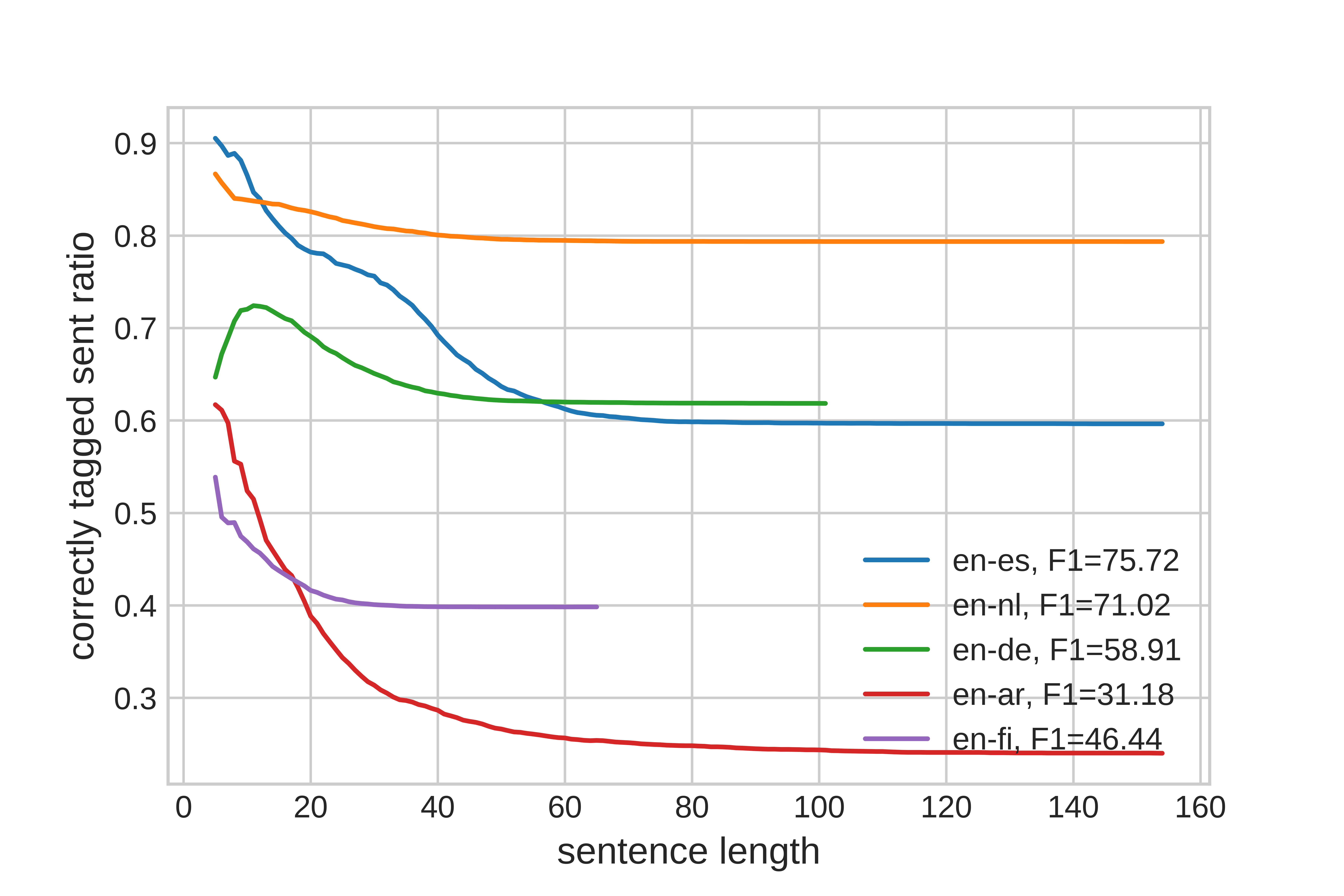}
  \caption{Sentence length vs. correctly tagged target words.}
  \label{fig:sent_dist:1}
\end{figure}

\subsection{Augmented Fine-tuning} 

\begin{figure*}[t!]
  \subcaptionbox{Mono-lingual embeddings
  \label{tsne-plot:1}}
  {\includegraphics[scale=.4]{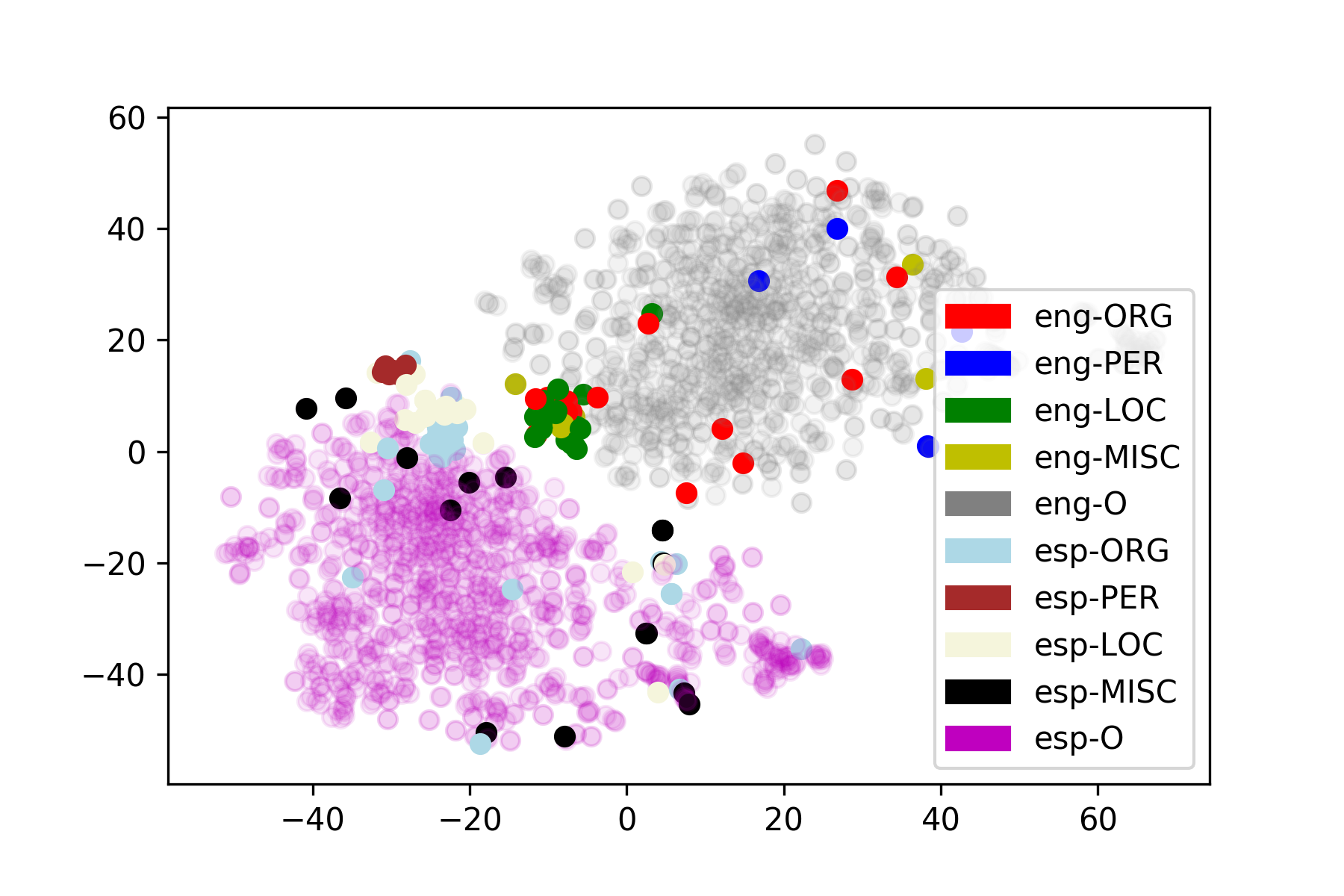}}
  \subcaptionbox{Cross-lingual embeddings \label{tsne-plot:2}}
  {\includegraphics[scale=.4]{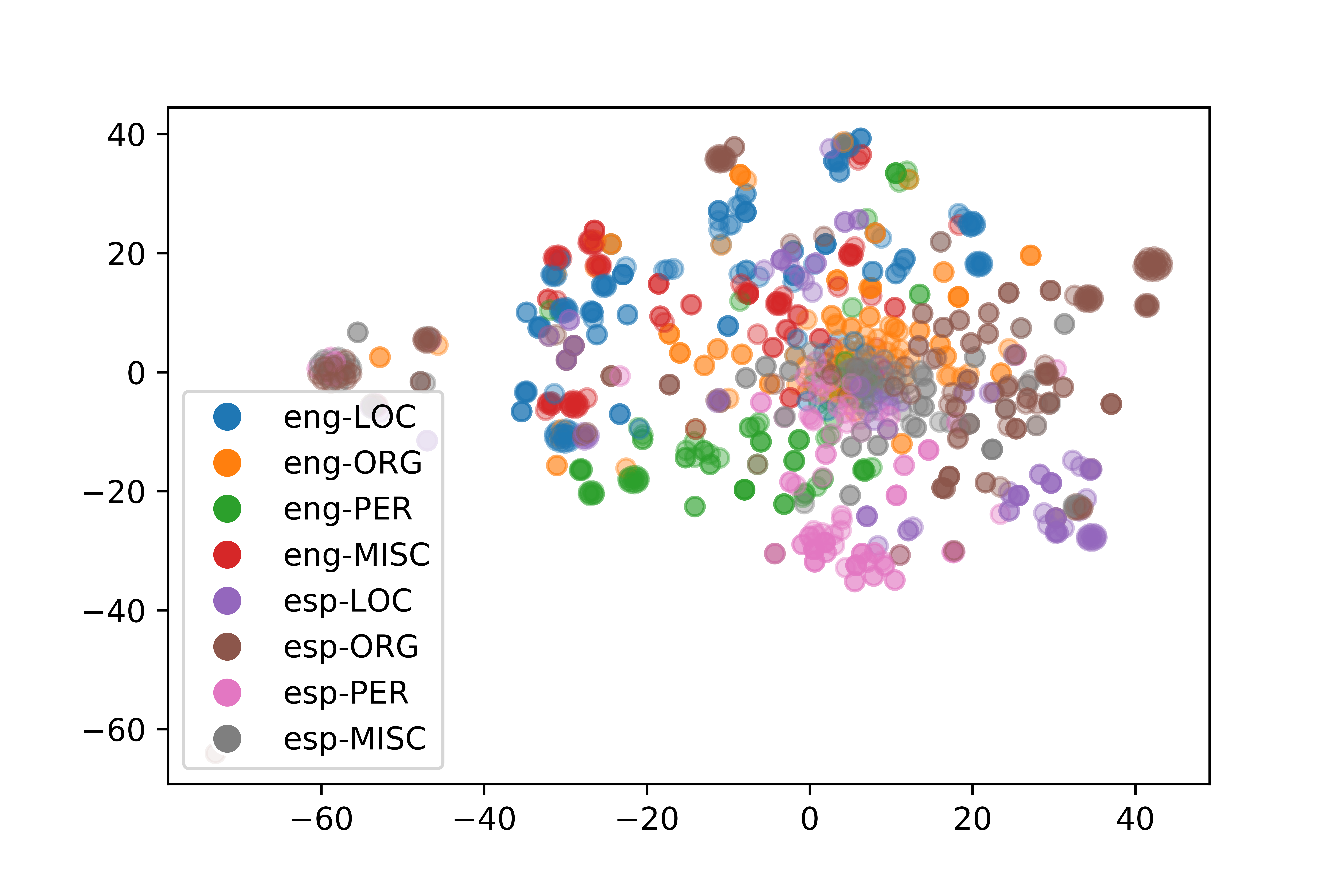}}
  \subcaptionbox{Common encoder output distribution \label{tsne-plot:3}}
  {\includegraphics[scale=.4]{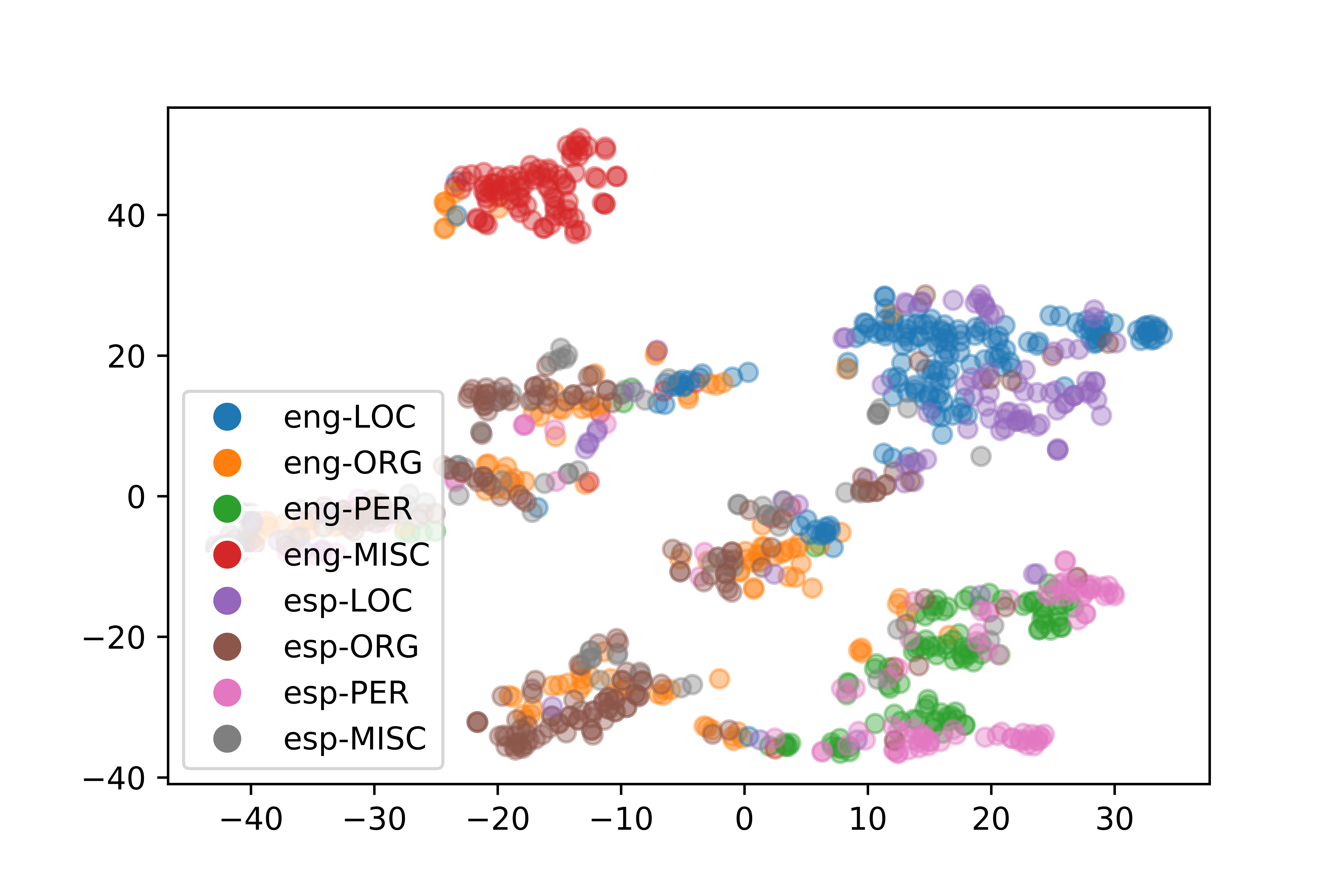}}
  \caption{t-SNE plot of NER tagged embeddings of two languages with 1000 samples: (a) Mono-lingual embeddings (fasttext), (b) Cross-lingual embeddings after word-level adversarial training, (c) Embeddings from our common encoder.}
  \label{fig:tsne}
\end{figure*}

The word-level adversarial training gives a mapping of the words independently. However, NER is a sequence labeling task, and the word order varies from one language to another. Besides, the word-level cross-lingual mapping process does not consider any task information (NER tags); it is simply a word translation model. As a result, the mappings may still lack alignments based on the NER tags. {This can be seen in Figure \ref{tsne-plot:2}, where the words are mapped to their translations but not clustered according to their NER tags.}

To learn target language ordering information in the target encoder and simultaneously transfer the NER knowledge from the source model, we propose a novel \emph{augmented fine-tuning} method, which works in three steps. 

\paragraph{(i) Source model pretraining through weight sharing.} We first train an NER model on the source where we have supervision. Our goal is to use this source model to generate pseudo NER labels for the target language sentences in the second step. Therefore, we train the model on the \emph{mapped} representation of the source words. Formally, we optimize: 
\begin{eqnarray}
{\sum_{i=1}^{P} \Ls^i_s(\theta_s|\mW_{s \rightarrow t})}
\label{eqn:mappsrc}
\end{eqnarray}
\noindent where $\Ls^i_s$ is the CRF classification loss in Equation \ref{eqn:crfloss} with $P$ being the number of training samples in the source.

The word order in the target language generally differs from the source. To make the model more effective on target sentences, we promote order invariant features in the source encoder by binding the parameters of the forward and backward layers of the character bi-LSTM and word bi-LSTM. Later in our experiments we show its effectiveness. Sharing also reduces the number of parameters and helps to achieve better generalization across languages \cite{lample2018phrase}. We will refer to this pretrained model as the \emph{mapped source model} or simply \emph{source model} parameterized by $\theta_s$. 

\paragraph{(ii) Generating pseudo target labels.} Since our source model is already trained in a cross-lingual space, it can directly be applied to infer the NER tags for the target sentences. As shown in Figure \ref{tsne-plot:2}, the word-level mapping provides good initial alignments that can be used to produce pseudo training samples in the target language to bootstrap training.

However, since the source model initially does not have any knowledge about the target language word order, it may generate noisy labels as the length of the target sentence increases. For example, Figure \ref{fig:sent_dist:1} shows the ratio of correctly tagged target words for different sentence lengths in different language pairs. We notice that the noise ratio is less for shorter sentences and it increases upto a point as the length increases. To effectively train our models with the pseudo target labels, we adopt a stochastic selection method based on sentence length. In particular, we randomly select a length threshold $l$ from a uniform distribution $\Us(min, max)$, where $min$ and $max$ are the minimum and maximum (target) sentence lengths respectively, and then we train our models only on sentences that have a maximum of $l$ words; see Algorithm \ref{alg:training}. This length restricted stochastic training schedule enables the model to tackle the learning-inference gap between short and long sentences.

\paragraph{(iii) Joint training with feature augmentation.} We train our target NER model jointly with the  source model with feature augmentation. For each batch from the source, we optimize our source model as before (Equation \ref{eqn:mappsrc}). For each target batch with pseudo labels, we jointly train the source and the target model, and the features from the source encoder are augmented with the features from the target encoder (see Figure \ref{fig:model}). The overall loss function of our model is:
\begin{eqnarray}
 \Ls (\theta_s, \theta_t) =  \underbrace{\sum_{i=1}^{P} \Ls^i_s(\theta_s|\mW_{s \rightarrow t})}_\text{source batch}  + \underbrace{\sum_{j=1}^{Q} \Ls^j_t(\theta_s)}_\text{target batch} + \underbrace{\sum_{j=1}^{Q} \Ls^j_t(\theta_t)}_\text{target batch}
\label{eqn:joint}
\end{eqnarray}
\noindent where $Q$ is the number of target samples considered for training. This joint training with augmented features ensures that the target model does not overfit on the (potentially) noisy target samples. In a way, the source model guides the target one. Algorithm \ref{alg:training} provides the pseudocode of our training method. Fig.  \ref{tsne-plot:3} shows a sample output distribution of our common encoder. We can see that the representations are now well clustered based on the NER tags.

\begin{algorithm2e}[t!]
\footnotesize
\SetKwInOut{Input}{Input}\SetKwInOut{Output}{Output}
\SetAlgoNoLine
\Input{Data $\Ds_S = \{x_i, y_i\}_{i=1}^P$, $\Ds_T = \{x_j\}_{j=1}^Q$, Monolingual Embeddings $\mE_s$ and $\mE_t$.}

\tcp{Word-level adversarial mapping}
1. \Repeat {w\_steps}{
        \Repeat {n\_disc\_steps}{
            i) Sample batches $b_s \sim \mE_s$ and $b_t \sim \mE_t$ \\
            ii) Update $\theta_D$ on disc. loss $\mathcal{L}_{D}(\theta_{D}|\mW)$ for $b_s$ and $b_t$\\
        }
        Sample batches $b_s \sim \mE_s$ and $b_t \sim \mE_t$ \\
        Update $\mW$ on adv. loss $\mathcal{L}_{\text{adv}}(\mW|\theta_{D})$ for $b_s$ and $b_t$\\
    }
\tcp{Source model pre-training}
2. \Repeat {n\_steps}{
        i) Sample a batch of sentences $b_s \sim \Ds_S$\\
        ii) Update $\theta_s$ on CRF classification loss $\Ls_s(\theta_s)$ for $b_s$\\
    }
\tcp{Augmented fine-tuning}
3. Sample a length-threshold $l$ from $\Us(min, max)$ \\

4. Use $\theta_s$ to infer on $\Ds_T$ to create a dataset $\Ds_T^l = \{x_j, \hat{y}_j\}_{j=1}^{Q_l}$\\

5. \Repeat {convergence}{ 
        \Repeat {n\_steps}{
            i) Sample a batch of sentences $b_s \sim \Ds_S$ and $b_t \sim \Ds_T^l$  \\
            ii) Update $\theta_s$ on CRF loss $\Ls_s(\theta_s)$ for $b_s$ and $b_t$ \\
            iii) Update $\theta_t$ on CRF loss $\Ls_t(\theta_t)$ for $b_t$
            
        }
        Sample a length-threshold $l$ from $\Us(min, max)$ \\
        Create a target dataset $\Ds_T^l = \{x_j,
		\hat{y}_j\}_{j=1}^{Q_l}$ using $\theta_s$ \\
}
\caption{\small Augmented fine-tuning for x-lingual NER}
\label{alg:training}
\end{algorithm2e}

\section{Experimental Settings}
\label{sec:exper}

\paragraph{Dataset} 

We experiment with five different target languages --- Spanish, Dutch, German, Arabic and Finnish. The source language is always English, for which we have sentences tagged with NER classes. The data for English is from the CoNLL-2003 shared task for NER \cite{Sang2003IntroductionTT}, while the data for Spanish and Dutch is from the CoNLL-2002 shared task for NER \cite{Sang2002IntroductionTT}. We collected the Finnish NER dataset from \cite{Ruokolainen_2019}\footnote{Available from https://github.com/mpsilfve/finer-data} and refactored a few tags. For Arabic, we use \emph{AQMAR Arabic Wikipedia Named Entity Corpus} \cite{AQMAR}.\footnote{http://www.cs.cmu.edu/~ark/ArabicNER/} The corpus contains 28 annotated Wikipedia articles. We randomly take 20\% of the sentences from each article to create development and test sets.\footnote{Both Arabic and Finnish dataset splits can be found at http://github.com//ntunlp/Zero-Shot-Cross-Lingual-NER} The NER data is tagged in the IOB1 format. Following the standard practice, we convert it to IOB2 to facilitate evaluation.  We train and validate our model in the IOBES format, which is more expressive, for all languages except Arabic.  Table \ref{table:datastats} presents some basic statistics of the datasets used in our experiments.

\begin{table}[t!]
\begin{center}
\resizebox{0.6\linewidth}{!}{
\begin{tabular}{l|ccc}
\toprule 
Language & Train & Dev. & Test \\
\midrule 
English   & 14041 & 3250 & 3453\\
Spanish	  & 8323 & 1915  & 1517\\
Dutch 	  & 15519 & 2821 & 5076\\
German    & 12152 & 2867 & 3005\\
Arabic    & 2166 & 267 & 254\\
Finnish    & 13497 & 986 & 3512\\
\bottomrule
\end{tabular}}
\caption{Training, Test and Development splits for different datasets. We exclude document start tags (\emph{DOCSTART}).} 
\label{table:datastats}
\end{center}
\end{table}

\paragraph{Compared Models}

We experiment with different baselines and variants of our model as described below.

\begin{itemize}[itemsep=0.0em]

    \item \textbf{Source-Mono:} We train an NER model on the source language with source word embeddings and apply it to the target language with target embeddings, which can be pre-trained or randomly initialized. This model does not use any cross-lingual information.
    
    \item \textbf{Cross-Word:} We project source and target word embeddings to a common space using the unsupervised mapper ($\mW_{s \rightarrow t}$ or $\mW_{t \rightarrow s}$). This model uses word-level cross-lingual information learned from adversarial training and the Procrustes-CSLS refinement procedure.

    \item \textbf{Cross-Shared:} This model is the same as \textbf{Cross-Word}, but the weights of the forward and backward LSTM cells are shared to encourage order invariance in the model.  

    \item \textbf{Cross-Augmented:} This is our full cross-lingual model trained with source labels and target pseudo-labels generated by the pretrained model and the model itself.
    
\end{itemize}

\paragraph{Model Settings}

We only use sentences with a maximum length of 250 words for training on the source language data. We use FastText embeddings \cite{grave2018learning}, which are trained on Common Crawl and Wikipedia, and SGD  with a gradient clipping of $5.0$ to train the model. We found that the learning rate was crucial for training, and used a decaying rate to scale it down after every epoch. In particular, the learning rate was set to $\max(\frac{{lr}_0}{1+decay*epoch}, 0.0001)$. The initial learning rate of ${lr}_0 = 0.1$ and $decay=0.01$ worked well with a dropout rate of $0.5$. We trained the model for $30$ epochs while using a batch size of $16$, and evaluated the model after every $150$ batches. The sizes of the character embeddings and char-LSTM hidden states were set to 25. Our word LSTM's hidden size was set to $100$. {The details of the hyperparameters are given in our Github repository.\footnote{https://github.com/ntunlp/Zero-Shot-Cross-Lingual-NER} We conducted all the experiments in Table \ref{table:weak_baseline} and Table \ref{table:results}  five (5) times, and report the mean, standard deviation and maximum value.}

\section{Results}

\subsection{Monolingual Results }

\begin{table}[t!]
\centering
\resizebox{0.9\linewidth}{!}{
\begin{tabular}{l|ccccc}
\toprule 
                  & Emb. type & Emb. dim & $F_1$ score\\
\midrule 
\textbf{English} \\
\cite{lampleNER} & random & 100 & 83.63 \\
\cite{lampleNER} & skip-ngram, no-char & 100 & 90.20 \\
\cite{lampleNER} & skip-ngram & 100 & 90.94 \\
Our & glove & 200 & 91.05$\Mypm$0.37 \\
Our & fasttext & 300 & 89.77$\Mypm$0.19 \\
\midrule
\textbf{Spanish} \\
\cite{lampleNER} & skip-ngram & 64 &  85.75 \\
\cite{cross-ling-cmu}	    & glove &  300 &  86.26 $\Mypm$0.40 \\
Our 	  & fasttext &  300 &  84.71$\Mypm$0.06  \\
\midrule
\textbf{Dutch} \\
\cite{lampleNER} & skip-ngram   & 64     & 81.74  \\
\cite{cross-ling-cmu}	    & glove        & 300    & 86.40$\Mypm$.17\\
Our 	                    & fasttext     & 300    & 85.16$\Mypm$0.21\\
\midrule
\textbf{German} \\
\cite{lampleNER} & skip-gram--no-char & 64 & 75.06\\
\cite{lampleNER} & skip-gram & 64 & 78.76 \\
 \cite{cross-ling-cmu} & glove & 200 & 78.16 $\Mypm$ 0.45 \\
Our         & fasttext   & 300     & 78.14 $\Mypm$ 0.32 \\
\midrule
\textbf{Arabic} \\
Our & fasttext   & 300     & 75.49$\Mypm$.53  \\
\midrule
\textbf{Finnish} \\
Our & fasttext   & 300     & 84.21$\Mypm$0.13  \\
\bottomrule
\end{tabular}
}

\caption{\textbf{Monolingual} NER results in the supervised setting.}
\label{table:monolingualresults}
\end{table}

 In Table \ref{table:monolingualresults}, we show the effect of different embeddings on the NER task.  We observe that character embeddings contribute very little towards learning the monolingual NER task. Though the monolingual model performs better with GloVe embeddings (Pennington et al. \citeyear{pennington-socher-manning:2014:EMNLP2014}), adversarial training performs better with FastText \cite{bojanowski2017enriching}, so we use FastText embeddings for all of our experiments.

\paragraph{Source-Mono}
In Table \ref{table:weak_baseline}, we show how the source base models perform when they are directly applied to the target language. We can see that the model only learns when character embeddings (shared) are used. Random word embeddings provide better results than monolingual word embeddings.

\begin{table}[t!]
\centering
\resizebox{1\linewidth}{!}{
\begin{tabular}{l|cccc}
\toprule 
Model & Language pair & $F_1^{*}$ score & $F_1$ score\\
\midrule 
\textbf{Mono-lingual word-emb} \\
Wrd-LSTM-CRF & en-\{es,nl,de,ar,fi\} & x & x \\
\midrule 
Ch-LSTM-Wrd-LSTM-CRF  & en-es & 33.66$\Mypm$0.90 & 26.76 $\Mypm$ 1.45  \\
Ch-LSTM-Wrd-LSTM-CRF & en-nl & 25.692 $\Mypm$ 1.75 & 20.94 $\Mypm$ 0.74 \\
Ch-LSTM-Wrd-LSTM-CRF & en-de & 12.54 $\Mypm$ 3.07 & 8.34 $\Mypm$ 1.43 \\
Ch-LSTM-Wrd-LSTM-CRF & en-ar & x & x
\\
Ch-LSTM-Wrd-LSTM-CRF & en-fi & 25.05 $\Mypm$ 0.54 & 22.44 $\Mypm$ 2.23 \\
\midrule 
\midrule 
\textbf{Random word-emb}  &  &  &  \\
Wrd-LSTM-CRF & en-\{es,nl,de,ar,fi\} & x  & x \\
\midrule 
Ch-LSTM-Wrd-LSTM-CRF & en-es & 36.87 $\Mypm$ 2.46& 32.61 $\Mypm$1.71\\
Ch-LSTM-Wrd-LSTM-CRF  & en-nl & 32.47 $\Mypm$0.92 & 24.74 $\Mypm$ 0.48 \\
Ch-LSTM-Wrd-LSTM-CRF & en-de & 14.70 $\Mypm$ 0.35& 11.51 $\Mypm$0.71\\
Ch-LSTM-Wrd-LSTM-CRF  & en-ar & x & x \\
Ch-LSTM-Wrd-LSTM-CRF  & en-fi & 26.05 $\Mypm$ 0.44 & 17.36 $\Mypm$ 3.34\\
\bottomrule
\end{tabular}
}

\caption{\label{table:weak_baseline} 
Results for monolingual models applied to target language NER task. `x' means the model fails to learn anything. $F_1^{*}$ and $F_1$ scores are calculated by tuning on the development datasets of the target and source, respectively.  }
\end{table}

\begin{table*}[t!]
\centering
\resizebox{0.98\linewidth}{!}{
\begin{tabular}{l|cccccccc}
\toprule 
Model &  Emb. prj. &   $F_1^{\spadesuit}$ (tuned on tgt-dev) & $F_1^{\spadesuit}$-max & $F_1^{\diamondsuit}$ (tuned on src-dev) & $F_1^{\diamondsuit}$-max & $F_1^{\clubsuit}$ (tuned on tgt-test) & $F_1^{\clubsuit}$-max & \# of params \\
\midrule 
Cross-Word & en $\rightarrow$ es & 68.63 $\Mypm$ 1.49 & 70.62 & 64.79 $\Mypm$ 1.68 & 67.42 & 68.90 $\Mypm$ 1.10 & 70.62 & 342906 ($\sim$13$\downarrow$)\\
(No char LSTM) & en $\rightarrow$ nl & 65.01  $\Mypm$ 0.53 & 65.73 & 64.28 $\Mypm$  0.71 & 65.05 & 65.86 $\Mypm$ 0.29 & 66.25 & 342906 ($\sim$13$\downarrow$)\\
 & en $\rightarrow$ de & 58.76 $\Mypm$ 0.70 & 59.7 & 57.12 $\Mypm$ 0.53 & 58.15 & 59.11 $\Mypm$ 0.37 & 59.7 & 342906 ($\sim$13$\downarrow$)\\
& en $\leftarrow$ ar & 29.81 $\Mypm$ 1.01 & 31.18 & 24.79 $\Mypm$ 0.65 & 25.46 & 30.74 $\Mypm$ 0.71 & 31.18 & 341890 ($\sim$14$\downarrow$)\\
& en $\leftarrow$ fi & 28.77 $\Mypm$ 1.19 & 30.01 & 26.55 $\Mypm$ 0.61 & 27.71 & 29.99 $\Mypm$ 0.36 & 30.44 & 342906 ($\sim$13$\downarrow$)\\

\midrule 
\multirow{2}{*}{Cross-Word} & en $\rightarrow$ es & 72.66 $\Mypm$ 0.39 & 73.19 & 70.49 $\Mypm$ 1.34 & 72.82  & 73.62 $\Mypm$ 0.70 & 74.76 & 395581 (=1x)\\
 & en $\rightarrow$ nl & 70.31 $\Mypm$ 1.01 & 71.5 &  69.24 $\Mypm$ 1.32 & 70.98 & 71.22 $\Mypm$  0.41 & 71.71 & 395881 (=1x)\\
 & en $\rightarrow$ de &  45.20 $\Mypm$ 2.78 & 48.94 & 30.99 $\Mypm$ 1.08 & 32.82 & 46.10 $\Mypm$ 1.68 & 48.94 & 395756 (=1x)\\
 & en $\leftarrow$ ar & 21.39 $\Mypm$ 1.85 & 24.6 & 11.84 $\Mypm$ 3.69 & 15.36 & 23.37 $\Mypm$ 1.40 & 24.77 & 396215 (=1x)\\
 & en $\leftarrow$ fi & 47.84 $\Mypm$ 1.12 & 49.53 & 44.90 $\Mypm$ 1.26 & 46.09 & 48.15 $\Mypm$ 0.88 & 49.53 & 395356 (=1x)\\
\midrule 
\midrule 
\multirow{2}{*}{Cross-Shared} & en $\rightarrow$ es & 74.39 $\Mypm$ 0.94 & 75.72 & 71.97 $\Mypm$ 0.85 & 72.54 & 74.91 $\Mypm$ 0.81 & 75.72 & 210081 ($\sim$47$\downarrow$)\\
  & en $\rightarrow$ nl & 71.02  $\Mypm$ 1.20 & 72.89 &  68.85 $\Mypm$ 1.87 & 70.69 & 71.62 $\Mypm$ 0.89 & 72.89 & 210381 ($\sim$47$\downarrow$)\\
 & en $\rightarrow$ de & 58.91  $\Mypm$ 1.03 & 60.35 & 56.20 $\Mypm$ 1.38 & 57.62 & 59.52 $\Mypm$  0.62 & 60.35 & 182506 ($\sim$54$\downarrow$)\\
 & en $\leftarrow$ ar & 28.28 $\Mypm$ 1.61 & 29.82 & 23.32 $\Mypm$ 0.76 & 24.35 & 29.89 $\Mypm$ 0.49 & 30.72 & 181490 ($\sim$54$\downarrow$)\\
  & en $\leftarrow$ fi & 48.04 $\Mypm$ 1.40 & 49.3 & 44.36 $\Mypm$ 2.52 & 48.37 & 49.31 $\Mypm$  0.69 & 50.13 & 209856 ($\sim$47$\downarrow$)\\
\midrule 
\midrule 
\midrule 
\multirow{2}{*}{\textbf{Cross-Augmented}} & en $\rightarrow$ es & 75.93 $\Mypm$ 0.81 & 77.03 & 72.36 $\Mypm$ 1.17 & 73.7 & 76.82 $\Mypm$ 0.84 & 77.81 & 661281 ($\sim$67$\uparrow$)\\
 & en $\rightarrow$ nl & 74.61 $\Mypm$1.24 & 76.43 & 69.43 $\Mypm$ 2.43 & 72.03 & 75.47 $\Mypm$ 1.25 & 77.45 & 661581 ($\sim$67$\uparrow$)\\
 & en $\rightarrow$ de & 65.24  $\Mypm$ 0.56 & 65.83 & 59.45 $\Mypm$  2.56 & 62.61 & 65.76 $\Mypm$ 0.41 & 66.02 & 636356  ($\sim$61$\uparrow$)\\
 & en $\leftarrow$ ar & 36.91  $\Mypm$ 2.74 & 40.36 & 27.12 $\Mypm$ 3.00 & 31.84 & 38.02 $\Mypm$ 2.41 & 41.63 & 797215  ($\sim$101$\uparrow$)\\
  & en $\leftarrow$ fi & 53.77  $\Mypm$ 1.54 & 56.05 & 45.69 $\Mypm$ 2.61 &50.67 & 54.42 $\Mypm$ 1.33 & 56.54 & 661056  ($\sim$67$\uparrow$)\\
\bottomrule
\end{tabular}
}
\caption{\label{table:results}\textbf{Cross-lingual} results for \textbf{English $\rightarrow$ Spanish}, \textbf{English $\rightarrow$ Dutch}, \textbf{English $\rightarrow$ German}, \textbf{English $\rightarrow$ Finnish} and \textbf{English $\rightarrow$ Arabic} with respect to different settings. We pick the best performing model amongst the \textbf{Cross-Word (No char LSTM)}, \textbf{Cross-Word} and \textbf{Cross-Shared} models. Using this model as the base, we train the \textbf{Cross Augmented} model.}

\end{table*}

\begin{table*}[h!]
\centering
\resizebox{1\linewidth}{!}{
\begin{tabular}{l|ccccccc}
\toprule 

Model &  Method & Word Emb. & \multicolumn{4}{c}{Lang. Pair} \\
      &         &      & en $\rightarrow$ es & en $\rightarrow$ nl & en $\rightarrow$ de & en $\rightarrow$ ar &  en $\rightarrow$ fi\\
\midrule
\textbf{with cross-lingual resources} & &   &    &  &  \\
Tackstrom et al. (\citeyear{Tackstrom:2012:CWC:2382029.2382096}) & Wiki article induction, parallel corpus & -  &   59.30   &  58.40 &  40.40 & - & - \\
\cite{Nothman:2013:LMN:2405838.2405915} & Word cluster features & -  &   60.55   &  61.60 &  48.10 & - & - \\
\cite{Tsai2016CrossLingualNE} & Feature based methods & -  &   61.0   &  64.00 &  55.80 & - & - \\ 
\cite{DBLP:journals/corr/NiDF17} & parallel corpus, dict & polyglot emb.  &   65.10   &  65.40 &  58.50 & - & -\\
\cite{Mayhew2017CheapTF} & Cheap Translation, multi-lingual & -  &  65.95   & 66.50 &  59.11 & - & - \\
\cite{Mayhew2017CheapTF} & Cheap Translation, english-only & -  &  51.82  & 53.94 &  50.96 & - & -\\
\midrule 
\textbf{without cross-lingual resources} & &   &    &  & \\
 \cite{cross-ling-cmu}  & Translate (train on translated src) & fasttext/MUSE, glove & 71.03 $\Mypm$ 0.44 & 71.25 $\Mypm$ 0.79  & 56.90 $\Mypm$ 0.76 & - & -\\
 \cite{Rahimi_NER} & Ranking and Retraining & fasttext/MUSE & 71.8 & 67.6  &  \textbf{59.1} & - & -\\
 \cite{Chen_NER} & MAN-MoE+CharCNN, multi-lingual  & fasttext/MUSE &  71.0  & 70.9 &  56.7 & - & -\\
 \cite{Chen_NER} & MAN-MoE+CharCNN, multi-lingual  & fasttext/UMWE &  \textbf{73.5}  & \textbf{72.4} &  56.0 & - & -\\
\midrule
\textbf{Our method} &  &  &  &  & \\
Cross-Shared & Common space proj (tgt$\rightarrow$ src) & fasttext/MUSE & 74.39 $\Mypm$ .94  & 71.02 $\Mypm$ 1.20 & 58.91 $\Mypm$ 1.03 & 28.28 $\Mypm$ 1.61 & 48.04 $\Mypm$ 1.40\\
Cross-Augmented & adaptation to tgt lang & fasttext/MUSE & \textbf{75.93} $\Mypm$ \textbf{0.81}  & \textbf{74.61} $\Mypm$ \textbf{1.24} & \textbf{65.24} $\Mypm$ \textbf{0.56} & \textbf{36.91} $\Mypm$ \textbf{2.74} & \textbf{53.77}  $\Mypm$ \textbf{1.54} \\
\bottomrule
\end{tabular}
}

\caption{Comparison of \textbf{Cross-lingual} NER results.}
\label{table:cmu-compare}
\end{table*}

\subsection{Cross-lingual Results}

\paragraph{Word-level Mapping}

For all language pairs except En-Ar and En-Fi, projecting word embeddings from the source to the target language achieves the best results. For En-Ar, we could not get reasonable results for source-to-target projection, which is an issue, as discussed by \citet{Non_adv_word_trans}.\footnote{See https://github.com/ntunlp/Zero-Shot-Cross-Lingual-NER for detailed results.}

\paragraph{Baseline Results}

From Table \ref{table:results} we can see that the \textbf{Cross-Word} model with character LSTM performs significantly better than the monolingual model (Source-Mono, Table \ref{table:monolingualresults}) for all languages.

\paragraph{Our Main Results}

The Cross-Shared model, in which the weights of the forward and backward LSTM cells are shared, gives us $1.73$ and $0.71$ absolute F1 score increments for the English to Spanish and Dutch  language pairs respectively, over the \textbf{Cross-Word} (with/without character) model  (Table \ref{table:results}). This already achieves a SOTA result by an absolute F1 score of $+0.89$ for the English-Spanish language pair.

Our Cross-Augmented model, (Tables \ref{table:results}-\ref{table:cmu-compare}) that performs adaptation from source to target language, achieves SOTA performance for all language pairs. It improves over the previous SOTA by $2.43$, $2.21$, $6.14$ and $5.73$ F1 for the English to Spanish, Dutch, German and Finnish language pairs respectively, even outperforming multi-lingual models.  Our model also outperforms the models that use cross-lingual resources for all languages - including German, which has not been the case in previous works. We also show the effectiveness of our model by reporting results on a ``proxy" low-resource\footnote{In the Wikipedia dump as of September 2019, Arabic/English size ratio is 891/16384 (in MB)=.0543 (~5.5\% of en) } dataset (Arabic), where there is no improvement using the \textbf{Cross-Shared} model, but a gain of $+7.1$ F1 using the \textbf{Cross-Augmented} method.

\section{Analysis}

\paragraph{Char embeddings}
Contrary to the monolingual case, we find that pretrained source character embeddings make a significant contribution towards transferring NER knowledge in the cross-lingual task, if the two languages have similar morphological features (en-es, en-nl, en-fi). For Arabic (does not share characters with English), the character embeddings only seem to work as noise. However, in case of German, there is a similar noise effect despite the shared characters. Presumably, this is because of the differences in the capitalisation patterns, since German capitalises all nouns.

\paragraph{Embedding distribution}
In the cross lingual model, the baseline results improve significantly. \ref{tsne-plot:1} and \ref{tsne-plot:2} show the distributions of the pairs of monolingual and cross-lingual embeddings. As the two languages do not share (Fig \ref{tsne-plot:1}) any space in their distribution, it is impossible for the model to learn anything. Monolingual embeddings also hamper training; random embeddings increase the transfer score (Table \ref{table:weak_baseline}), but the model performs poorly with random embeddings for monolingual training (Table \ref{table:monolingualresults}). However, the result improves in \ref{table:weak_baseline}. This suggests that we need to search for a better common space for both languages; thus, we perform cross-lingual projection by adversarial training.

\paragraph{Shared LSTM cell}
In order to get better sequence invariance, we experimented with shared weights in forward and backward LSTM cells. This comes from the idea of \textbf{learning less to transfer more}. For Spanish and Dutch, this leads to significant improvements in results along with a 47\% reduction in parameters. For German and Finnish there is no significant difference, but the number of parameters are reduced by ~54\% and ~47\%.  However, for Arabic, there is a drop in the results, probably because of significant word-order differences with the source language (English).

\paragraph{Effect of Sentence Length}
One of our main assumptions is that pseudo-labels can reduce the entropy of the model \cite{Entropy_min_bengio}. Sentence length is a good feature for finding better pseudo-labels. However, this comes with a cost. To study the effect of sentence length while training the Cross-Augmented model, we perform experiments with sentences of lengths varying from 30 to 150. Figure \ref{fig:sent_dist:1} shows that as the sentence length increases, the ratio of correctly tagged sentences reduces. But if we only train the model on short sentences, the model will overfit on the short sentences of the target language data. Our main model addresses this issue by adding a teacher model and randomly sampling sentence lengths from a uniform distribution.

\paragraph{Source vs. Target NER distribution}
 We report the results of our model tuned on both target and source development data. We see that the model tuned on target development data performs better than the model tuned on source dev data. The results of the source dev data tuned model should be considered as the results under a purely unsupervised setting. 
 These results highlight the differences between the source and target NER distributions. Tuning on the target dev data therefore plays a significant role in the results obtained in cross-lingual NER research thus far. 
 We also tried tuning the model with target test data. Here also we observe a gap between the results. To report stable results, the standard practice should be to report the results of multiple experiments with their standard deviations. Until now, to our knowledge, the only other paper to follow this has been \citet{cross-ling-cmu}.

\section{Related Work}

 \citet{lampleNER} proposed an LSTM-CRF model for NER, which passes a hierarchical bi-LSTM encoding to a CRF layer to encourage global consistency of the NER tags. 
 This model achieved impressive results for EN, NL, DE and ES despite not using any explicit feature engineering or manual gazetteers. We extend this base model to a cross-lingual named entity recognizer for a target language using annotated data for a source language and only monolingual, unannotated data for the target. 

 Mayhew et al. (\citeyear{Mayhew2017CheapTF}) use a dictionary and co-occurrence probabilities to generate word and phrase based translations of the source data into a target data and then transfer the labels; although the translation quality is poor, the words/phrases and most of the relevant context is preserved, and they are able to achieve good results using a combination of orthographic and Wikifier (Tsai et al. \citeyear{Tsai2016CrossLingualNE})  features. Ni et al. (\citeyear{DBLP:journals/corr/NiDF17}) use weak supervision for cross-lingual NER where they do annotation projection to get target labels and project word embeddings from the target language to the source language. Finally, Yang et al. (\citeyear{ICLR2017:TransferLearning}) used a hierarchical recurrent network for semi-supervised cross-language transfer learning, where the source and the target language share the same character embeddings. 
  Xie et al. (\citeyear{cross-ling-cmu}) are the first to propose a neural-based model for cross-lingual NER using the \cite{lampleNER} model, with the addition of a self-attention layer on top of word representation, and validate the model based on target side development dataset.

\section{Acknowledgement}
We thank Jiateng Xie, Guillaume Lample and Emma Strubell for sharing their code and embeddings, and  for their helpful replies on Github issues and e-mail. Also thanks to Tasnim Mohiuddin for a useful discussion on the  hyperparameters of the Word Translation model.

\section{Conclusions and Future Work}
In this paper, we contribute a detailed definition of the problem of cross-lingual NER, thus providing a structure to the research to come hereafter. We also propose a new method for cross-lingual NER that generalizes well by weight-sharing and iteratively adapting to the target language domain, achieving SOTA in the process across languages from different language families.
In future work, we want to explore pre-trained language models for cross-lingual NER transfer.

\bibliographystyle{aaai.bst}
\bibliography{AAAI-BariM.7577}

\end{document}